\documentclass[sigconf]{acmart}

\copyrightyear{2024}
\acmYear{2024}
\setcopyright{rightsretained}
\acmConference[WWW '24 Companion]{Companion Proceedings of the ACM Web Conference 2024}{May 13--17, 2024}{Singapore, Singapore}
\acmBooktitle{Companion Proceedings of the ACM Web Conference 2024 (WWW '24 Companion), May 13--17, 2024, Singapore, Singapore}\acmDOI{10.1145/3589335.3651470}
\acmISBN{979-8-4007-0172-6/24/05}

\usepackage{amsmath}
\usepackage{amsthm}
\usepackage{amsfonts}
\usepackage{nicematrix}
\usepackage{booktabs}
\usepackage{array}
\usepackage{graphicx}
\usepackage{multirow}
\usepackage{tikz}
\usepackage{enumitem}
\usepackage{subcaption}
\usepackage{arydshln}

\NiceMatrixOptions
  {
    custom-line = 
     {
       letter = : ,
       command = dashedline , 
       ccommand = cdashedline ,
       tikz = dashed
     }
  }

\usepackage{amsmath,amsfonts,bm}

\def\eqref#1{equation~\ref{#1}}

\def\1{\bm{1}}
\newcommand{\train}{\mathcal{D}}

\def\vx{{\bm{x}}}

\DeclareMathAlphabet{\mathsfit}{\encodingdefault}{\sfdefault}{m}{sl}
\SetMathAlphabet{\mathsfit}{bold}{\encodingdefault}{\sfdefault}{bx}{n}

\def\gC{{\mathcal{C}}}

\def\gL{{\mathcal{L}}}

\def\gR{{\mathcal{R}}}

\def\gW{{\mathcal{W}}}

\def\gY{{\mathcal{Y}}}

\def\sR{{\mathbb{R}}}

\newtheoremstyle{mystyle}
  {3pt}
  {3pt}
  {}
  {}
  {\bfseries}
  {.}
  {.5em}
  {\thmname{#1}\thmnumber{ #2}:\thmnote{ #3}}
  
\theoremstyle{mystyle}
\newtheorem{definition}{Definition}

\newcommand{\norm}[1]{\left \lVert #1 \right \rVert}

\makeatletter
\gdef\@copyrightpermission{
  \begin{minipage}{0.3\columnwidth}
   \href{https://creativecommons.org/licenses/by/4.0/}{\includegraphics[width=0.90\textwidth]{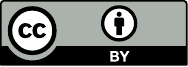}}
  \end{minipage}\hfill
  \begin{minipage}{0.7\columnwidth}
   \href{https://creativecommons.org/licenses/by/4.0/}{This work is licensed under a Creative Commons Attribution International 4.0 License.}
  \end{minipage}
  \vspace{5pt}
}
\makeatother

\begin{document}

\title[Revisiting Chemical Reaction Yield Prediction from an Imbalanced Regression Perspective]{Are we Making Much Progress? Revisiting Chemical Reaction Yield Prediction from an Imbalanced Regression Perspective}

\author{Yihong Ma}
\affiliation{
  \institution{University of Notre Dame}
  \country{}
}
\email{yma5@nd.edu}

\author{Xiaobao Huang}
\affiliation{
  \institution{University of Notre Dame}
  \country{}
}
\email{xhuang2@nd.edu}

\author{Bozhao Nan}
\affiliation{
  \institution{University of Notre Dame}
  \country{}
}
\email{bnan@nd.edu}

\author{Nuno Moniz}
\affiliation{
  \institution{University of Notre Dame}
  \country{}
}
\email{nuno.moniz@nd.edu}

\author{Xiangliang Zhang}
\affiliation{
  \institution{University of Notre Dame}
  \country{}
}
\email{xzhang33@nd.edu}

\author{Olaf Wiest}
\affiliation{
  \institution{University of Notre Dame}
  \country{}
}
\email{owiest@nd.edu}

\author{Nitesh V. Chawla}
\affiliation{
  \institution{University of Notre Dame}
  \country{}
}
\email{nchawla@nd.edu}

\renewcommand{\shortauthors}{Yihong Ma et al.}

\begin{abstract}
The yield of a chemical reaction quantifies the percentage of the target product formed in relation to the reactants consumed during the chemical reaction. Accurate yield prediction can guide chemists toward selecting high-yield reactions during synthesis planning, offering valuable insights before dedicating time and resources to wet lab experiments. While recent advancements in yield prediction have led to overall performance improvement across the entire yield range, an open challenge remains in enhancing predictions for high-yield reactions, which are of greater concern to chemists. In this paper, we argue that \emph{the performance gap in high-yield predictions results from the imbalanced distribution of real-world data skewed towards low-yield reactions, often due to unreacted starting materials and inherent ambiguities in the reaction processes}. Despite this data imbalance, existing yield prediction methods continue to treat different yield ranges equally, assuming a balanced training distribution. Through extensive experiments on three real-world yield prediction datasets, we emphasize the urgent need to reframe reaction yield prediction as an imbalanced regression problem. Finally, we demonstrate that incorporating simple cost-sensitive re-weighting methods can significantly enhance the performance of yield prediction models on underrepresented high-yield regions.
\end{abstract}



\maketitle

\section{Introduction}

Recent advancements in machine learning have introduced a paradigm shift in the field of computational chemistry \cite{guo2022graph}. These breakthroughs have led to a diverse array of machine learning models that now play critical roles in assisting chemists across a broad spectrum of tasks, including but not limited to retrosynthesis, product prediction, and drug discovery.
Among this multifaceted landscape, the prediction of reaction yields \cite{ahneman2018predicting, schwaller2021prediction, saebi2023use, voinarovska2023yield, yarish2023advancing} emerges as an issue of paramount importance in the domain of synthesis planning, where complex molecules are synthesized through a sequence of reaction steps. Based on the empirical categorization, yields above 67\% are classified as high yields and those below 33\% are classified as low yields \cite{yarish2023advancing}.  In this context, the occurrence of a low-yield reaction within this sequence can drastically impact the feasibility and overall efficiency of the synthesis process. As a result, chemists often prioritize the accurate prediction of high-yield reactions.

While the introduction of numerous yield prediction models has indeed showcased improved performance across the entire yield range, the challenge of effectively enhancing performance for high-yield reactions remains an open problem \cite{maloney2023negative}. In real-world scenarios, yield data often exhibits a highly imbalanced distribution, with high yield values being much rarer than lower ones, despite their greater importance to chemists in synthesis planning. 
In this paper, we argue that \emph{the increased difficulty in predicting high-yield reactions stems from its limited availability of data samples, often due to unreacted starting materials and inherent ambiguities in the reaction processes}. Despite the presence of such data imbalance, existing yield prediction methods continue to treat different yield ranges equally with the false assumption of a balanced data distribution.

To gain a deeper insight into the field's actual progress, we conduct extensive experiments to benchmark six state-of-the-art yield prediction methods on three real-world datasets. Surprisingly, the results become less impressive than claimed when we take data imbalance into account. We discover that \emph{the overall good performance across the entire yield spectrum primarily results from enhancing performance in areas with sufficient data, typically the low-yield range, while overlooking the significant performance gap in underrepresented high-yield regions}. This finding has motivated us to revisit reaction yield prediction and reformulate it as an imbalanced regression problem, a well-established topic in machine learning.

Unlike imbalanced classification, reaction yield prediction involves regression rather than classification, and there has been limited exploration of addressing data imbalance in the regression context \cite{ribeiro2020imbalanced}. Most prior research on imbalanced regression has directly adapted the SMOTE algorithm \cite{chawla2002smote} to regression settings \cite{branco2017smogn, torgo2013smote}. However, the continuous nature of target labels in regression tasks makes these adaptations less practical. A more intuitive solution is to apply cost-sensitive re-weighting strategies \cite{gong2022ranksim,lin2017focal,yang2021delving} that can be seamlessly combined with various regression models. We demonstrate that incorporating these simple methods
can significantly enhance the performance of existing yield prediction models on underrepresented high-yield regions without sacrificing the overall performance too much. We believe these findings have the potential to redirect the future research direction in reaction yield prediction, benefiting both chemistry and machine learning communities. In summary, the contributions of this paper include:
\begin{itemize}[leftmargin=*]
    \item We are the first to introduce the novel concept of reformulating reaction yield prediction as an imbalanced regression problem.
    \item We conduct comprehensive experiments on three real-world yield prediction datasets to uncover and understand the limitations of existing models when predicting high-yield reactions.
    \item We demonstrate that incorporating cost-sensitive re-weighting methods into existing yield prediction models can lead to significant performance improvements on high-yield reactions.
\end{itemize}

\section{the examination of existing yield prediction methods}
In this section, we begin by introducing the definitions of reaction yield prediction and imbalanced regression. We then proceed to evaluate six yield prediction methods on three real-world datasets.

\vspace{-0.05in}
\subsection{Preliminaries}

\begin{table*}[t]
\resizebox{0.9\textwidth}{!}{
\begin{NiceTabular}{|c|c|cccc|cccc|cccc|} 

\toprule

\multirow{2.4}{*}{\textbf{Dataset}} & \multirow{2.4}{*}{\textbf{Method}} & \multicolumn{4}{c}{\textbf{MAE} $\downarrow$} & \multicolumn{4}{c}{\textbf{RMSE} $\downarrow$} & \multicolumn{4}{c}{\textbf{G-Mean} $\downarrow$}\\ 
\cmidrule{3-14}
& & All & Many & Med. & Few & All & Many & Med. & Few & All & Many & Med. & Few \\

\midrule

\multirow{6}{*}{B-H} & RF & 5.5\footnotesize{$\pm 0.2$} & 4.4\footnotesize{$\pm 0.2$} & 6.0\footnotesize{$\pm 0.2$} & 6.7\footnotesize{$\pm 0.6$} & 8.1\footnotesize{$\pm 0.3$} & 7.4\footnotesize{$\pm 0.3$} & 8.4\footnotesize{$\pm 0.4$} & 9.1\footnotesize{$\pm 0.8$} & 2.8\footnotesize{$\pm 0.1$} & 1.8\footnotesize{$\pm 0.1$} & 3.6\footnotesize{$\pm 0.2$} & 4.1\footnotesize{$\pm 0.4$} \\
& XGBoost & 4.7\footnotesize{$\pm 0.2$} & 3.8\footnotesize{$\pm 0.2$} & 5.2\footnotesize{$\pm 0.2$} & 5.7\footnotesize{$\pm 0.5$} & 6.9\footnotesize{$\pm 0.3$} & 6.1\footnotesize{$\pm 0.4$} & 7.1\footnotesize{$\pm 0.5$} & 7.8\footnotesize{$\pm 0.7$} & 2.7\footnotesize{$\pm 0.1$} & 2.0\footnotesize{$\pm 0.1$} & 3.2\footnotesize{$\pm 0.1$} & 3.4\footnotesize{$\pm 0.3$} \\
& SVM & 14.6\footnotesize{$\pm 0.3$} & 12.9\footnotesize{$\pm 0.5$} & 12.9\footnotesize{$\pm 0.7$} & 26.1\footnotesize{$\pm 1.1$} & 18.5\footnotesize{$\pm 0.4$} & 15.8\footnotesize{$\pm 0.5$} & 17.2\footnotesize{$\pm 0.8$} & 28.6\footnotesize{$\pm 0.9$} & 9.2\footnotesize{$\pm 0.3$} & 8.7\footnotesize{$\pm 0.5$} & 7.7\footnotesize{$\pm 0.6$} & 22.2\footnotesize{$\pm 1.6$} \\
& MLP & 4.5\footnotesize{$\pm 0.2$} & 3.4\footnotesize{$\pm 0.3$} & 5.3\footnotesize{$\pm 0.2$} & 5.3\footnotesize{$\pm 0.4$} & 7.0\footnotesize{$\pm 0.5$} & 6.0\footnotesize{$\pm 1.0$} & 7.6\footnotesize{$\pm 0.6$} & 7.3\footnotesize{$\pm 0.6$} & 2.5\footnotesize{$\pm 0.1$} & 1.8\footnotesize{$\pm 0.1$} & 3.1\footnotesize{$\pm 0.1$} & 3.3\footnotesize{$\pm 0.3$} \\
& YieldGNN & 8.7\footnotesize{$\pm 7.9$} & 8.6\footnotesize{$\pm 9.5$} & 7.6\footnotesize{$\pm 4.6$} & 12.8\footnotesize{$\pm 14.5$} & 11.1\footnotesize{$\pm 8.7$} & 10.6\footnotesize{$\pm 9.1$} & 10.1\footnotesize{$\pm 5.9$} & 14.7\footnotesize{$\pm 14.7$} & 3.0\footnotesize{$\pm 0.4$} & 2.3\footnotesize{$\pm 0.5$} & 3.5\footnotesize{$\pm 0.4$} & 3.9\footnotesize{$\pm 0.5$} \\
& Yield-BERT & 5.5\footnotesize{$\pm 0.3$} & 3.8\footnotesize{$\pm 0.3$} & 6.6\footnotesize{$\pm 0.4$} & 6.6\footnotesize{$\pm 0.5$} & 8.4\footnotesize{$\pm 0.4$} & 7.1	\footnotesize{$\pm 1.4$} & 9.1\footnotesize{$\pm 0.6$} & 8.9\footnotesize{$\pm 0.8$} & 3.2\footnotesize{$\pm 0.1$} & 2.2\footnotesize{$\pm 0.1$} & 4.1\footnotesize{$\pm 0.4$} & 4.1\footnotesize{$\pm 0.5$} \\
\midrule
\multicolumn{2}{c}{\textbf{Avg. Ranking}} & - & \textbf{1.2} & \textbf{1.8} & \textbf{3.0} & - & \textbf{1.2} & \textbf{2.2} & \textbf{2.7} & - & \textbf{1.2} & \textbf{1.8} & \textbf{3.0} \\ 

\midrule

\multirow{6}{*}{S-M} & RF & 8.0\footnotesize{$\pm 0.2$} & 6.1\footnotesize{$\pm 0.4$} & 8.8\footnotesize{$\pm 0.3$} & 12.8\footnotesize{$\pm 6.1$} & 11.8\footnotesize{$\pm 0.4$} & 11.2\footnotesize{$\pm 0.9$} & 12.0\footnotesize{$\pm 0.3$} & 15.9\footnotesize{$\pm 7.6$} & 4.0\footnotesize{$\pm 0.2$} & 2.3\footnotesize{$\pm 0.1$} & 5.1\footnotesize{$\pm 0.3$} & 9.8\footnotesize{$\pm 3.8$} \\
& XGBoost & 7.2\footnotesize{$\pm 0.2$} & 6.1\footnotesize{$\pm 0.4$} & 7.7\footnotesize{$\pm 0.2$} & 6.8\footnotesize{$\pm 4.7$} & 10.5\footnotesize{$\pm 0.2$} & 10.1\footnotesize{$\pm 0.8$} & 10.7\footnotesize{$\pm 0.2$} & 9.7\footnotesize{$\pm 6.3$} & 4.1\footnotesize{$\pm 0.1$} & 3.1\footnotesize{$\pm 0.2$} & 4.6\footnotesize{$\pm 0.1$} & 3.9\footnotesize{$\pm 2.7$} \\
& SVM & 16.5\footnotesize{$\pm 0.2$} & 14.8\footnotesize{$\pm 0.5$} & 17.1\footnotesize{$\pm 0.2$} & 31.6\footnotesize{$\pm 6.0$} & 20.3\footnotesize{$\pm 0.2$} & 18.7\footnotesize{$\pm 0.6$} & 20.8\footnotesize{$\pm 0.3$} & 33.0\footnotesize{$\pm 6.1$} & 11.1\footnotesize{$\pm 0.2$} & 9.6\footnotesize{$\pm 0.3$} & 11.7\footnotesize{$\pm 0.3$} & 30.3\footnotesize{$\pm 6.0$} \\
& MLP & 7.5\footnotesize{$\pm 0.3$} & 6.1\footnotesize{$\pm 0.5$} & 8.1\footnotesize{$\pm 0.4$} & 8.5\footnotesize{$\pm 6.8$} & 11.1\footnotesize{$\pm 0.4$} & 10.7\footnotesize{$\pm 0.9$} & 11.2\footnotesize{$\pm 0.5$} & 12.3\footnotesize{$\pm 9.3$} & 4.1\footnotesize{$\pm 0.2$} & 2.9\footnotesize{$\pm 0.2$} & 4.8\footnotesize{$\pm 0.3$} & 4.5\footnotesize{$\pm 3.5$} \\
& YieldGNN & 9.1\footnotesize{$\pm 1.2$} & 7.2\footnotesize{$\pm 0.9$} & 9.9\footnotesize{$\pm 1.9$} & 10.8\footnotesize{$\pm 7.3$} & 12.6\footnotesize{$\pm 1.5$} & 10.8\footnotesize{$\pm 1.4$} & 13.2\footnotesize{$\pm 2.2$} & 13.5\footnotesize{$\pm 8.2$} & 5.4\footnotesize{$\pm 0.6$} & 4.1\footnotesize{$\pm 0.5$} & 6.1\footnotesize{$\pm 1.2$} & 8.0\footnotesize{$\pm 6.9$} \\
& Yield-BERT & 9.1\footnotesize{$\pm 0.4$} & 6.3\footnotesize{$\pm 0.4$} & 10.2\footnotesize{$\pm 0.5$} & 5.1\footnotesize{$\pm 5.4$} & 13.3\footnotesize{$\pm 0.6$} & 11.2\footnotesize{$\pm 0.9$} & 14.1\footnotesize{$\pm 6.3$} & 7.7\footnotesize{$\pm 7.6$} & 5.0\footnotesize{$\pm 0.2$} & 3.3\footnotesize{$\pm 0.2$} & 6.0\footnotesize{$\pm 0.3$} & 2.3\footnotesize{$\pm 2.7$} \\
\midrule
\multicolumn{2}{c}{\textbf{Avg. Ranking}} & - & \textbf{1.2} & \textbf{2.3} & \textbf{2.5} & - & \textbf{1.3} & \textbf{2.3} & \textbf{2.3} & - & \textbf{1.2} & \textbf{2.5} & \textbf{2.3} \\ 

\midrule

\multirow{6}{*}{AZ} & RF & 20.3\footnotesize{$\pm 0.8$} & 19.1\footnotesize{$\pm 0.7$} & 23.6\footnotesize{$\pm 2.5$} & 32.0\footnotesize{$\pm 4.3$} & 25.2\footnotesize{$\pm 0.9$} & 23.6\footnotesize{$\pm 0.8$} & 29.8\footnotesize{$\pm 3.0$} & 36.4\footnotesize{$\pm 3.9$} & 13.8\footnotesize{$\pm 0.8$} & 13.1\footnotesize{$\pm 0.6$} & 16.0\footnotesize{$\pm 3.0$} & 26.0\footnotesize{$\pm 6.4$} \\
& XGBoost & 20.6\footnotesize{$\pm 1.0$} & 19.5\footnotesize{$\pm 1.0$} & 23.9\footnotesize{$\pm 3.4$} & 30.4\footnotesize{$\pm 6.1$} & 27.2\footnotesize{$\pm 1.3$} & 25.6\footnotesize{$\pm 1.3$} & 31.9\footnotesize{$\pm 3.6$} & 37.9\footnotesize{$\pm 6.0$} & 11.8\footnotesize{$\pm 1.0$} & 11.3\footnotesize{$\pm 1.1$} & 14.0\footnotesize{$\pm 3.6$} & 19.7\footnotesize{$\pm 6.5$} \\
& SVM & 25.0\footnotesize{$\pm 0.8$} & 23.5\footnotesize{$\pm 0.7$} & 28.0\footnotesize{$\pm 2.4$} & 41.7\footnotesize{$\pm 3.9$} & 28.9\footnotesize{$\pm 0.9$} & 27.1\footnotesize{$\pm 0.7$} & 32.5\footnotesize{$\pm 2.4$} & 43.9\footnotesize{$\pm 3.8$} & 18.6\footnotesize{$\pm 0.9$} & 17.5\footnotesize{$\pm 0.7$} & 20.8\footnotesize{$\pm 2.9$} & 37.6\footnotesize{$\pm 5.8$} \\
& MLP & 22.1\footnotesize{$\pm 1.7$} & 21.5\footnotesize{$\pm 1.3$} & 25.0\footnotesize{$\pm 5.0$} & 26.0\footnotesize{$\pm 6.9$} & 29.7\footnotesize{$\pm 2.1$} & 28.9\footnotesize{$\pm 1.9$} & 33.4\footnotesize{$\pm 4.9$} & 32.8\footnotesize{$\pm 7.4$} & 12.0\footnotesize{$\pm 1.2$} & 11.6\footnotesize{$\pm 0.9$} & 14.6\footnotesize{$\pm 5.0$} & 15.6\footnotesize{$\pm 6.7$} \\
& YieldGNN & 22.5\footnotesize{$\pm 0.8$} & 21.5\footnotesize{$\pm 0.6$} & 25.5\footnotesize{$\pm 3.3$} & 33.4\footnotesize{$\pm 6.7$} & 28.0\footnotesize{$\pm 1.0$} & 26.5\footnotesize{$\pm 1.0$} & 32.0\footnotesize{$\pm 3.7$} & 38.6\footnotesize{$\pm 6.0$} & 15.2\footnotesize{$\pm 1.1$} & 14.6\footnotesize{$\pm 0.9$} & 17.0\footnotesize{$\pm 3.1$} & 25.7\footnotesize{$\pm 9.2$} \\
& Yield-BERT & 25.6\footnotesize{$\pm 2.5$} & 24.3\footnotesize{$\pm 2.6$} & 28.7\footnotesize{$\pm 3.8$} & 38.4\footnotesize{$\pm 10.1$} & 32.5\footnotesize{$\pm 3.0$} & 30.7\footnotesize{$\pm 3.0$} & 37.5\footnotesize{$\pm 3.6$} & 45.2\footnotesize{$\pm 9.9$} & 15.7\footnotesize{$\pm 1.5$} & 15.1\footnotesize{$\pm 1.7$} & 18.0\footnotesize{$\pm 4.7$} & 26.2\footnotesize{$\pm 10.8$} \\
\midrule
\multicolumn{2}{c}{\textbf{Avg. Ranking}} & - & \textbf{1.0} & \textbf{2.0} & \textbf{3.0} & - & \textbf{1.0} & \textbf{2.2} & \textbf{2.8} & - & \textbf{1.0} & \textbf{2.0} & \textbf{3.0} \\ 

\bottomrule
\end{NiceTabular}
}
\caption{Reaction yield prediction results on three real-world datasets. We report the average performance across 10 repetitions in all experiments and present the average rankings for the \emph{many-shot}, \emph{medium-shot}, and \emph{few-shot} regions, respectively.}
\vspace{-0.3in}
\label{tab:exp}
\end{table*}

\begin{definition}[Reaction yield prediction]
    Reaction yield prediction is a regression problem that predicts the yield value $y \in [0, 100]$ of a chemical reaction $rxn = (\gR, P)$ composed of multiple reactant molecules $\gR$ and a single product molecule $P$.
\end{definition}

\begin{definition}[Imbalanced regression]

    Let $\train = \{ (\vx_i, y_i) \}_{i=1}^{N}$ denote the training dataset of a regression problem, where $\vx_i \in \sR^d$ is the input feature vector and $y_i \in \sR$ is the label. We divide the label space $\gY$ into $K$ disjoint bins with equal intervals, i.e., $[b_0, b_1), [b_1, b_2), \dots, [b_{K-1}, b_K)$. Let $\gC_k$ be the set of data samples in the $k$-th bin with $k \in \{ 1, 2, \dots, B \}$. The data imbalance occurs when the label distribution is highly skewed, i.e., $\frac{\max_k |\gC_k|}{\min_k |\gC_k|} \gg 1$.
\end{definition}

\begin{figure}[t]
    \centering
    \includegraphics[width=0.45\textwidth]{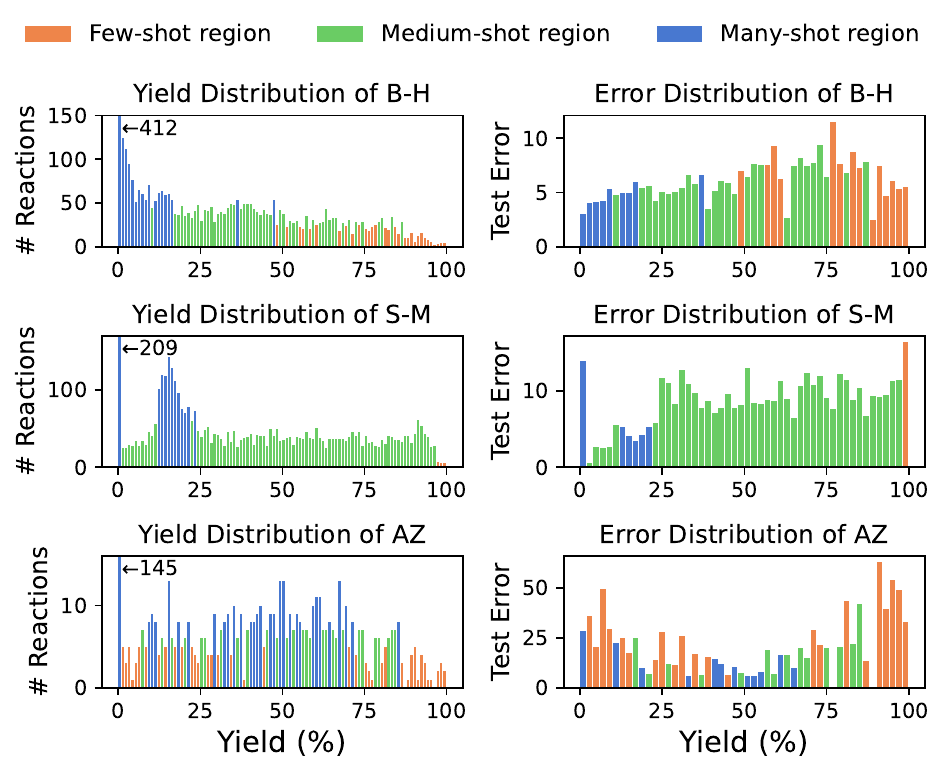}
    \vspace{-0.15in}
    \caption{A comparison between yield distributions (left) and test error distributions (right) on three real-world datasets.}
    \label{fig:dist}
    \vspace{-0.2in}
\end{figure}

\vspace{-0.1in}
\subsection{Evaluation Settings for Yield Prediction}

\subsubsection{Datasets}
We use three real-world datasets for predicting reaction yields, sourced from either high-throughput experimentation (HTE) or electronic laboratory notebooks (ELN). Following prior research on imbalanced regression \cite{yang2021delving, gong2022ranksim, liu2023semi}, we categorize the bins within the target yield space into three disjoint subsets: \emph{many-shot} (bins with over $\#_\text{upper}$ reactions), \emph{medium-shot} (bins with $\#_\text{lower}$ to $\#_\text{upper}$ reactions), and \emph{few-shot} (bins with fewer than $\#_\text{lower}$ reactions) regions, based on their respective numbers of training samples. For all three datasets, we set the bin size $|b_k - b_{k-1}|$ to 1. The left section of Figure \ref{fig:dist} visualizes the division of these regions.
\begin{itemize}[leftmargin=*]
    \item B-H \cite{ahneman2018predicting}: It comprises 3,955 Buchward-Hartwig reactions from HTE, with the number of reactions per bin varying between 1 and 412. Here, $\#_\text{lower}$ is set to 25, and $\#_\text{upper}$ is set to 50.
    \item S-M \cite{perera2018platform}: It consists of 5,760 Suzuki-Miyaura reactions from HTE, with the number of reactions per bin ranging from 1 to 209. Here, $\#_\text{lower}$ is set to 20, and $\#_\text{upper}$ is set to 65.
    \item AZ \cite{saebi2023use}: It includes 750 Buchward-Hartwig reactions from ELN at AstraZeneca, with the number of reactions per bin ranging from 0 to 145. Here, $\#_\text{lower}$ is set to 3, and $\#_\text{upper}$ is set to 5.
\end{itemize}

\subsubsection{Yield prediction methods}
\begin{itemize}[leftmargin=*]
\item \textbf{Machine learning methods:} Random Forest (RF), XGBoost, Support Vector Machines (SVM);
\item \textbf{Deep learning methods:} Multi-layer Perceptron (MLP), YieldGNN \cite{saebi2023use}, Yield-BERT \cite{schwaller2021prediction}.
\end{itemize}

\vspace{-0.05in}
\subsubsection{Evaluation pipeline}
\label{subsubsec:eval}

We report yield prediction results on \emph{many-shot}, \emph{medium-shot}, and \emph{few-shot} regions as well as on the entire yield space (i.e., the \emph{all} region). In all three datasets, 70\% of the data is used for training and the remaining 30\% is reserved for testing. Our evaluation employs common yield prediction metrics: mean absolute error (MAE) and root mean square error (RMSE). Additionally, we also utilize the geometric mean of $L_1$ errors (G-Mean) as a supplementary metric. Lower values ($\downarrow$) of MAE, RMSE, and G-Mean indicate better yield prediction performance. 

\vspace{-0.05in}
\subsubsection{Implementation details}
For RF, XGBoost, SVM, and MLP, the input features include structural fingerprints (e.g., ECFP), chemical properties (e.g., NMR shifts, HOMO/LUMO energies, vibrations, dipole moments), and reaction-specific parameters (e.g., scale, volume, temperature). For Yield-BERT, the SMILES string of the reaction is used as input; an encoder based on a pre-trained BERT \cite{kenton2019bert} for SMILES is employed, with a k-Nearest Neighbors (kNN) regressor serving as the decoder. For YieldGNN, we construct graph structures to represent the molecules involved in the reaction; a GNN is used to encode the reaction, while an MLP is used to decode the reaction embedding into yield predictions. We employ the $L_1$ distance as the training loss $\gL$ in all experiments.

\vspace{-0.1in}
\subsection{Uncovering and Understanding the Performance Gap in High-Yield Predictions}
Figure \ref{fig:dist} provides an insightful comparison between yield distributions and test error distributions, both as functions of yield values. Regarding the yield distribution, we note that the \emph{few-shot} region predominantly comprises high-yield reactions, while the \emph{many-shot} region primarily consists of low-yield reactions. Specifically in the B-H and S-M datasets, 81\% and 100\% of the \emph{few-shot} reactions fall into the high-yield category, while 89\% and 100\% of the \emph{many-shot} reactions belong to the low-yield category. This finding establishes a clear connection between yield values and their distributions.

To quantify the impact of data imbalance on prediction errors, we compute the Pearson correlation coefficients between the testing error distribution and the training yield distribution. Across all three reaction yield prediction datasets, we consistently observe negative correlation coefficients, with values of -0.42, -0.28, and -0.07, respectively. Moreover, in the right part of Figure \ref{fig:dist}, it is evident that the \emph{few-shot} region (in orange) exhibits the largest test error, while the \emph{many-shot} region (in blue) demonstrates the smallest error. To complement this observation, we further evaluate the yield prediction performance of six state-of-the-art models using three metrics and report the average performance rankings for the three regions in Table \ref{tab:exp}. As a result, the average ranking indicates a decline in model performance as we transition from the \emph{many-shot} region to the \emph{medium-shot} and \emph{few-shot} regions.

Therefore, both observations from Figure \ref{fig:dist} and Table \ref{tab:exp} converge to the same conclusion: During the training process, low-yield values with a larger number of data samples tend to be learned better in comparison to those high yields with fewer samples. This highlights the demand for specialized machine learning techniques that can effectively address the challenge of data imbalance.

\section{mitigating data imbalance for better high-yield predictions}
In this section, we present two cost-sensitive re-weighting methods for imbalanced regression, which can be seamlessly integrated into the learning process of existing yield prediction models. We also provide evidence of their effectiveness in high-yield predictions.

\vspace{-0.05in}
\subsection{Cost-sensitive Re-weighting Methods}
The key idea is to assign a weight $w_i \in \gW$ to each training sample $(\vx_i, y_i)$, resulting in the following modified loss function $\gL'$:
\begin{equation}
    \gL' = \frac{1}{N} \sum_{i=1}^N w_i \gL(y_i, \hat{y}_i),
\end{equation}
where $\gL$ is a loss function for regression tasks, such as $L_1$ loss, MSE loss, and Huber loss. The distinctiveness of each re-weighting method arises from the various designs of training weights $\gW$.

\vspace{-0.05in}
\subsubsection{Focal loss}
The Focal loss \cite{lin2017focal} is a specialized loss function designed to address imbalanced learning problems. It assigns varying weights to individual training samples based on their prediction difficulties. The goal is to reduce the impact of easily predictable samples while amplifying the importance of challenging samples during the training process. The weight $w_i$ is defined as:
\begin{equation}
    w_i = \textsc{sigmoid}(\alpha \gL(y_i, \hat{y}_i))^\gamma
\end{equation}
where $\textsc{sigmoid}(\cdot)$ is the sigmoid function, and $\alpha$, $\gamma$ are hyper-parameters. This results in the scaling factor of each training sample ranging from 0 to 1, depending on the prediction error. Notably, when $\gamma = 0$, the Focal loss is equivalent to the original loss $\gL$. In all experiments, we set $\alpha$ to 0.2 and $\gamma$ to 1.

\vspace{-0.05in}
\subsubsection{Label distribution smoothing}
Label distribution smoothing (LDS) \cite{yang2021delving} assigns varying weights to training samples based on their label density. The objective is to mitigate the impact of redundant samples while accentuating the significance of sparsely represented samples during training. To account for the continuity of labels, a Gaussian kernel is employed to smooth the empirical label density distribution of the label space $\gY$. The weight $w_i$ for each training sample is defined as:
\begin{equation}
    w_i = \frac{1}{\int_\gY K(y_i, y) |\gC_k| dy},
\end{equation}
where $K(y, y') = \exp \left( -\frac{\norm{y - y'}^2}{2 \sigma^2} \right)$ represents a Gaussian kernel with kernel size $\ell$ and standard deviation $\sigma$, and $\gC_k$ denotes the set of training samples in the $k$-th bin where $y_i \in [b_{k-1}, b_k)$. In all experiments, we set $\ell$ to 5 and $\sigma$ to 2. 

\begin{table}[t]
\resizebox{0.45\textwidth}{!}{
\begin{NiceTabular}{|c|c|cc|cc|cc|} 

\toprule

\multirow{2.4}{*}{\textbf{Dataset}} & \multirow{2.4}{*}{\textbf{Method}} & \multicolumn{2}{c}{\textbf{MAE} $\downarrow$} & \multicolumn{2}{c}{\textbf{RMSE} $\downarrow$} & \multicolumn{2}{c}{\textbf{G-Mean} $\downarrow$} \\ 
\cmidrule{3-8}
& & All & Few & All & Few & All & Few \\

\midrule

\multirow{4}{*}{B-H} & \multicolumn{1}{l}{Vanilla} & \textbf{4.5}\footnotesize{$\pm 0.2$} & 5.3\footnotesize{$\pm 0.4$} & 7.0\footnotesize{$\pm 0.5$} & 7.3\footnotesize{$\pm 0.6$} & \textbf{2.5}\footnotesize{$\pm 0.1$} & 3.3\footnotesize{$\pm 0.3$}\\
& \multicolumn{1}{l}{+Focal} & \underline{4.6}\footnotesize{$\pm 0.2$} & \underline{5.0}\footnotesize{$\pm 0.3$} & \textbf{6.5}\footnotesize{$\pm 0.3$} & \underline{6.8}\footnotesize{$\pm 0.5$} & \underline{2.8}\footnotesize{$\pm 0.1$} & \underline{3.0}\footnotesize{$\pm 0.4$}\\
& \multicolumn{1}{l}{+LDS} & 4.8\footnotesize{$\pm 0.3$} & \textbf{4.6}\footnotesize{$\pm 0.5$} & 7.1\footnotesize{$\pm 0.7$} & \textbf{6.3}\footnotesize{$\pm 0.5$} & 2.8\footnotesize{$\pm 0.2$} & \textbf{2.8}\footnotesize{$\pm 0.4$}\\
& \multicolumn{1}{l}{+Focal+LDS} & 4.7\footnotesize{$\pm 0.2$} & 5.4\footnotesize{$\pm 0.4$} & \underline{6.7}\footnotesize{$\pm 0.2$} & 7.3\footnotesize{$\pm 0.5$} & \underline{2.8}\footnotesize{$\pm 0.1$} & 3.4\footnotesize{$\pm 0.4$}\\

\midrule

\multirow{4}{*}{S-M} & \multicolumn{1}{l}{Vanilla} & \textbf{7.5}\footnotesize{$\pm 0.3$} & 8.5\footnotesize{$\pm 6.8$} & \textbf{11.1}\footnotesize{$\pm 0.4$} & 12.3\footnotesize{$\pm 9.3$} & \textbf{4.1}\footnotesize{$\pm 0.2$} & 4.5\footnotesize{$\pm 3.5$}\\
& \multicolumn{1}{l}{+Focal} & 8.5\footnotesize{$\pm 0.1$} & 7.0\footnotesize{$\pm 2.3$} & 12.0\footnotesize{$\pm 0.3$} & 10.0\footnotesize{$\pm 3.6$} & 5.0\footnotesize{$\pm 0.1$} & \underline{3.5}\footnotesize{$\pm 1.1$}\\
& \multicolumn{1}{l}{+LDS} & \underline{8.0}\footnotesize{$\pm 0.3$} & \underline{6.1}\footnotesize{$\pm 2.8$} & \underline{11.}3\footnotesize{$\pm 0.4$} & \underline{8.1}\footnotesize{$\pm 3.8$} & \underline{4.7}\footnotesize{$\pm 0.2$} & 3.8\footnotesize{$\pm 1.7$}\\
& \multicolumn{1}{l}{+Focal+LDS} & 8.6\footnotesize{$\pm 0.3$} & \textbf{5.7}\footnotesize{$\pm 2.2$} & 12.0\footnotesize{$\pm 0.4$} & \textbf{7.8}\footnotesize{$\pm 3.2$} & 5.1\footnotesize{$\pm 0.1$} & \textbf{3.0}\footnotesize{$\pm 1.3$}\\

\midrule

\multirow{4}{*}{AZ} & \multicolumn{1}{l}{Vanilla} & \underline{22.1}\footnotesize{$\pm 1.7$} & 26.0\footnotesize{$\pm 6.9$} & 29.7\footnotesize{$\pm 2.1$} & \underline{32.8}\footnotesize{$\pm 7.4$} & \textbf{12.0}\footnotesize{$\pm 1.2$} & 15.6\footnotesize{$\pm 6.7$}\\
& \multicolumn{1}{l}{+Focal} & \textbf{22.0}\footnotesize{$\pm 1.0$} & 26.2\footnotesize{$\pm 4.4$} & 29.5\footnotesize{$\pm 1.3$} & 33.2\footnotesize{$\pm 5.0$} & 13.0\footnotesize{$\pm 0.7$} & 16.1\footnotesize{$\pm 5.9$}\\
& \multicolumn{1}{l}{+LDS} & 22.2\footnotesize{$\pm 1.3$} & \textbf{24.4}\footnotesize{$\pm 6.5$} & \textbf{29.1}\footnotesize{$\pm 1.6$} & \textbf{30.6}\footnotesize{$\pm 6.8$} & 12.9\footnotesize{$\pm 0.8$} & \underline{14.8}\footnotesize{$\pm 7.3$}\\
& \multicolumn{1}{l}{+Focal+LDS} & \textbf{22.0}\footnotesize{$\pm 1.0$} & \underline{25.7}\footnotesize{$\pm 5.9$} & \underline{29.3}\footnotesize{$\pm 1.2$} & 33.2\footnotesize{$\pm 7.6$} & \underline{12.5}\footnotesize{$\pm 0.8$} & \textbf{14.2}\footnotesize{$\pm 4.9$}\\

\bottomrule
\end{NiceTabular}
}
\caption{Comparison of reaction yield prediction performance with and without cost-sensitive re-weighting methods on \emph{all} and \emph{few-shot} regions. The best results are highlighted in bold, and the second-best results are underlined.}
\vspace{-0.3in}
\label{tab:exp2}
\end{table}

\subsection{Effectiveness in High-Yield Predictions}
Table \ref{tab:exp2} presents the experiment results on the same three yield prediction datasets, demonstrating the effectiveness of cost-sensitive re-weighting methods (i.e., Focal loss, LDS, and the combination of both) when integrated with existing yield prediction models. Across all three datasets, imbalanced regression methods consistently achieve the best results in all 9 combinations of evaluation metrics (MAE, RMSE, and G-Mean) in the \emph{few-shot} region. Meanwhile, the base model (``Vanilla'') without any imbalanced regression designs maintains its superiority in 6 out of 9 combinations in the \emph{all} region. This observation is understandable and aligns with the trade-off between prediction performance in underrepresented data regions and performance across the entire dataset. 

However, it's important to note that while there is a performance drop in the \emph{all} region, this drop is not significant compared to the substantial performance improvement observed in the \emph{few-shot} region. Specifically, the average performance drops are 6.7\%, 15.7\%, and 0.7\%, while the average performance improvements are 14.0\%, 34.3\%, and 7.3\% on each dataset, respectively. This observation underscores the effectiveness of incorporating simple imbalanced regression methods into existing yield prediction models as a plug-in module. Furthermore, it is important to highlight that by tailoring more sophisticated imbalanced regression techniques to yield prediction, we could anticipate a further performance improvement.

\section{Conclusion}
In this paper, we have highlighted a critical issue in reaction yield prediction – the prevalent focus on achieving superior performance across the entire yield spectrum, often at the expense of overlooking yield regions with limited training samples, particularly the high-yield areas, which are of greater concern to chemists. Through extensive experiments on three real-world datasets, we have emphasized the urgent need to reframe reaction yield prediction as an imbalanced regression problem. Moreover, we have demonstrated that by incorporating simple cost-sensitive re-weighting techniques, we can significantly enhance the performance of yield prediction models in underrepresented high-yield regions. 

\begin{acks}
This work was supported by National Science Foundation
through the NSF Center for Computer-Assisted Synthesis (C-CAS), under grant number CHE-2202693.
\end{acks}

\bibliographystyle{ACM-Reference-Format}
\bibliography{refs}


\begin{thebibliography}{17}


\ifx \showCODEN    \undefined \def \showCODEN     #1{\unskip}     \fi
\ifx \showDOI      \undefined \def \showDOI       #1{#1}\fi
\ifx \showISBNx    \undefined \def \showISBNx     #1{\unskip}     \fi
\ifx \showISBNxiii \undefined \def \showISBNxiii  #1{\unskip}     \fi
\ifx \showISSN     \undefined \def \showISSN      #1{\unskip}     \fi
\ifx \showLCCN     \undefined \def \showLCCN      #1{\unskip}     \fi
\ifx \shownote     \undefined \def \shownote      #1{#1}          \fi
\ifx \showarticletitle \undefined \def \showarticletitle #1{#1}   \fi
\ifx \showURL      \undefined \def \showURL       {\relax}        \fi
\providecommand\bibfield[2]{#2}
\providecommand\bibinfo[2]{#2}
\providecommand\natexlab[1]{#1}
\providecommand\showeprint[2][]{arXiv:#2}

\bibitem[Ahneman et~al\mbox{.}(2018)]%
        {ahneman2018predicting}
\bibfield{author}{\bibinfo{person}{Derek~T Ahneman}, \bibinfo{person}{Jes{\'u}s~G Estrada}, \bibinfo{person}{Shishi Lin}, \bibinfo{person}{Spencer~D Dreher}, {and} \bibinfo{person}{Abigail~G Doyle}.} \bibinfo{year}{2018}\natexlab{}.
\newblock \showarticletitle{Predicting reaction performance in C--N cross-coupling using machine learning}.
\newblock \bibinfo{journal}{\emph{Science}} (\bibinfo{year}{2018}).
\newblock


\bibitem[Branco et~al\mbox{.}(2017)]%
        {branco2017smogn}
\bibfield{author}{\bibinfo{person}{Paula Branco}, \bibinfo{person}{Lu{\'\i}s Torgo}, {and} \bibinfo{person}{Rita~P Ribeiro}.} \bibinfo{year}{2017}\natexlab{}.
\newblock \showarticletitle{SMOGN: a pre-processing approach for imbalanced regression}. In \bibinfo{booktitle}{\emph{First international workshop on learning with imbalanced domains: Theory and applications}}.
\newblock


\bibitem[Chawla et~al\mbox{.}(2002)]%
        {chawla2002smote}
\bibfield{author}{\bibinfo{person}{Nitesh~V Chawla}, \bibinfo{person}{Kevin~W Bowyer}, \bibinfo{person}{Lawrence~O Hall}, {and} \bibinfo{person}{W~Philip Kegelmeyer}.} \bibinfo{year}{2002}\natexlab{}.
\newblock \showarticletitle{SMOTE: synthetic minority over-sampling technique}.
\newblock \bibinfo{journal}{\emph{Journal of artificial intelligence research}} (\bibinfo{year}{2002}).
\newblock


\bibitem[Gong et~al\mbox{.}(2022)]%
        {gong2022ranksim}
\bibfield{author}{\bibinfo{person}{Yu Gong}, \bibinfo{person}{Greg Mori}, {and} \bibinfo{person}{Frederick Tung}.} \bibinfo{year}{2022}\natexlab{}.
\newblock \showarticletitle{{R}ank{S}im: Ranking Similarity Regularization for Deep Imbalanced Regression}. In \bibinfo{booktitle}{\emph{ICML}}.
\newblock


\bibitem[Guo et~al\mbox{.}(2023)]%
        {guo2022graph}
\bibfield{author}{\bibinfo{person}{Zhichun Guo}, \bibinfo{person}{Kehan Guo}, \bibinfo{person}{Bozhao Nan}, \bibinfo{person}{Yijun Tian}, \bibinfo{person}{Roshni~G Iyer}, \bibinfo{person}{Yihong Ma}, \bibinfo{person}{Olaf Wiest}, \bibinfo{person}{Xiangliang Zhang}, \bibinfo{person}{Wei Wang}, \bibinfo{person}{Chuxu Zhang}, {et~al\mbox{.}}} \bibinfo{year}{2023}\natexlab{}.
\newblock \showarticletitle{Graph-based molecular representation learning}. In \bibinfo{booktitle}{\emph{IJCAI}}.
\newblock


\bibitem[Kenton and Toutanova(2019)]%
        {kenton2019bert}
\bibfield{author}{\bibinfo{person}{Jacob Devlin Ming-Wei~Chang Kenton} {and} \bibinfo{person}{Lee~Kristina Toutanova}.} \bibinfo{year}{2019}\natexlab{}.
\newblock \showarticletitle{Bert: Pre-training of deep bidirectional transformers for language understanding}. In \bibinfo{booktitle}{\emph{NAACL}}.
\newblock


\bibitem[Lin et~al\mbox{.}(2017)]%
        {lin2017focal}
\bibfield{author}{\bibinfo{person}{Tsung-Yi Lin}, \bibinfo{person}{Priya Goyal}, \bibinfo{person}{Ross Girshick}, \bibinfo{person}{Kaiming He}, {and} \bibinfo{person}{Piotr Doll{\'a}r}.} \bibinfo{year}{2017}\natexlab{}.
\newblock \showarticletitle{Focal loss for dense object detection}. In \bibinfo{booktitle}{\emph{ICCV}}.
\newblock


\bibitem[Liu et~al\mbox{.}(2023)]%
        {liu2023semi}
\bibfield{author}{\bibinfo{person}{Gang Liu}, \bibinfo{person}{Tong Zhao}, \bibinfo{person}{Eric Inae}, \bibinfo{person}{Tengfei Luo}, {and} \bibinfo{person}{Meng Jiang}.} \bibinfo{year}{2023}\natexlab{}.
\newblock \showarticletitle{Semi-Supervised Graph Imbalanced Regression}. In \bibinfo{booktitle}{\emph{KDD}}.
\newblock


\bibitem[Maloney et~al\mbox{.}(2023)]%
        {maloney2023negative}
\bibfield{author}{\bibinfo{person}{Michael~P Maloney}, \bibinfo{person}{Connor~W Coley}, \bibinfo{person}{Samuel Genheden}, \bibinfo{person}{Nessa Carson}, \bibinfo{person}{Paul Helquist}, \bibinfo{person}{Per-Ola Norrby}, {and} \bibinfo{person}{Olaf Wiest}.} \bibinfo{year}{2023}\natexlab{}.
\newblock \bibinfo{title}{Negative Data in Data Sets for Machine Learning Training}.
\newblock
\newblock


\bibitem[Perera et~al\mbox{.}(2018)]%
        {perera2018platform}
\bibfield{author}{\bibinfo{person}{Damith Perera}, \bibinfo{person}{Joseph~W Tucker}, \bibinfo{person}{Shalini Brahmbhatt}, \bibinfo{person}{Christopher~J Helal}, \bibinfo{person}{Ashley Chong}, \bibinfo{person}{William Farrell}, \bibinfo{person}{Paul Richardson}, {and} \bibinfo{person}{Neal~W Sach}.} \bibinfo{year}{2018}\natexlab{}.
\newblock \showarticletitle{A platform for automated nanomole-scale reaction screening and micromole-scale synthesis in flow}.
\newblock \bibinfo{journal}{\emph{Science}} (\bibinfo{year}{2018}).
\newblock


\bibitem[Ribeiro and Moniz(2020)]%
        {ribeiro2020imbalanced}
\bibfield{author}{\bibinfo{person}{Rita~P Ribeiro} {and} \bibinfo{person}{Nuno Moniz}.} \bibinfo{year}{2020}\natexlab{}.
\newblock \showarticletitle{Imbalanced regression and extreme value prediction}.
\newblock \bibinfo{journal}{\emph{Machine Learning}} (\bibinfo{year}{2020}).
\newblock


\bibitem[Saebi et~al\mbox{.}(2023)]%
        {saebi2023use}
\bibfield{author}{\bibinfo{person}{Mandana Saebi}, \bibinfo{person}{Bozhao Nan}, \bibinfo{person}{John~E Herr}, \bibinfo{person}{Jessica Wahlers}, \bibinfo{person}{Zhichun Guo}, \bibinfo{person}{Andrzej~M Zura{\'n}ski}, \bibinfo{person}{Thierry Kogej}, \bibinfo{person}{Per-Ola Norrby}, \bibinfo{person}{Abigail~G Doyle}, \bibinfo{person}{Nitesh~V Chawla}, {et~al\mbox{.}}} \bibinfo{year}{2023}\natexlab{}.
\newblock \showarticletitle{On the use of real-world datasets for reaction yield prediction}.
\newblock \bibinfo{journal}{\emph{Chemical Science}} (\bibinfo{year}{2023}).
\newblock


\bibitem[Schwaller et~al\mbox{.}(2021)]%
        {schwaller2021prediction}
\bibfield{author}{\bibinfo{person}{Philippe Schwaller}, \bibinfo{person}{Alain~C Vaucher}, \bibinfo{person}{Teodoro Laino}, {and} \bibinfo{person}{Jean-Louis Reymond}.} \bibinfo{year}{2021}\natexlab{}.
\newblock \showarticletitle{Prediction of chemical reaction yields using deep learning}.
\newblock \bibinfo{journal}{\emph{Machine learning: science and technology}} (\bibinfo{year}{2021}).
\newblock


\bibitem[Torgo et~al\mbox{.}(2013)]%
        {torgo2013smote}
\bibfield{author}{\bibinfo{person}{Lu{\'\i}s Torgo}, \bibinfo{person}{Rita~P Ribeiro}, \bibinfo{person}{Bernhard Pfahringer}, {and} \bibinfo{person}{Paula Branco}.} \bibinfo{year}{2013}\natexlab{}.
\newblock \showarticletitle{Smote for regression}. In \bibinfo{booktitle}{\emph{Portuguese conference on artificial intelligence}}.
\newblock


\bibitem[Voinarovska et~al\mbox{.}(2023)]%
        {voinarovska2023yield}
\bibfield{author}{\bibinfo{person}{Varvara Voinarovska}, \bibinfo{person}{Mikhail Kabeshov}, \bibinfo{person}{Dmytro Dudenko}, \bibinfo{person}{Samuel Genheden}, {and} \bibinfo{person}{Igor~V Tetko}.} \bibinfo{year}{2023}\natexlab{}.
\newblock \showarticletitle{When yield prediction does not yield prediction: an overview of the current challenges}.
\newblock \bibinfo{journal}{\emph{Journal of Chemical Information and Modeling}} (\bibinfo{year}{2023}).
\newblock


\bibitem[Yang et~al\mbox{.}(2021)]%
        {yang2021delving}
\bibfield{author}{\bibinfo{person}{Yuzhe Yang}, \bibinfo{person}{Kaiwen Zha}, \bibinfo{person}{Yingcong Chen}, \bibinfo{person}{Hao Wang}, {and} \bibinfo{person}{Dina Katabi}.} \bibinfo{year}{2021}\natexlab{}.
\newblock \showarticletitle{Delving into deep imbalanced regression}. In \bibinfo{booktitle}{\emph{ICLR}}.
\newblock


\bibitem[Yarish et~al\mbox{.}(2023)]%
        {yarish2023advancing}
\bibfield{author}{\bibinfo{person}{Dzvenymyra Yarish}, \bibinfo{person}{Sofiya Garkot}, \bibinfo{person}{Oleksandr~O Grygorenko}, \bibinfo{person}{Dmytro~S Radchenko}, \bibinfo{person}{Yurii~S Moroz}, {and} \bibinfo{person}{Oleksandr Gurbych}.} \bibinfo{year}{2023}\natexlab{}.
\newblock \showarticletitle{Advancing molecular graphs with descriptors for the prediction of chemical reaction yields}.
\newblock \bibinfo{journal}{\emph{Journal of Computational Chemistry}} (\bibinfo{year}{2023}).
\newblock


\end{thebibliography}

\end{document}